\documentclass[pdflatex]{sn-jnl}

\jyear{2021}%

\theoremstyle{thmstyleone}%
\theoremstyle{thmstyletwo}%

\theoremstyle{thmstylethree}%

\begin{document}

\title[ ]{Transformer-based Time-to-Event Prediction for Chronic Kidney Disease Deterioration
}

\author[1]{\fnm{Moshe} \sur{Zisser}}

\author*[2,3]{\fnm{Dvir} \sur{Aran}}\email{dviraran@technion.ac.il}


\affil[1]{\orgdiv{Faculty of Data and Decision Sciences}, \orgname{Technion-Israel Institute of Technology}, \city{Haifa}, \country{Israel}}

\affil[2]{\orgdiv{Faculty of Biology}, \orgname{Technion-Israel Institute of Technology}, \city{Haifa}, \country{Israel}}

\affil[3]{\orgdiv{The Taub Faculty of Computer Science}, \orgname{Technion-Israel Institute of Technology}, \city{Haifa}, \country{Israel}}



\abstract{
Deep-learning techniques, particularly the transformer model, have shown great potential in enhancing the prediction performance of longitudinal health records. While previous methods have mainly focused on fixed-time risk prediction, time-to-event prediction (also known as survival analysis) is often more appropriate for clinical scenarios. Here, we present a novel deep-learning architecture we named STRAFE, a generalizable survival analysis transformer-based architecture for electronic health records. The performance of STRAFE was evaluated using a real-world claim dataset of over 130,000 individuals with stage 3 chronic kidney disease (CKD) and was found to outperform other time-to-event prediction algorithms in predicting the exact time of deterioration to stage 5. Additionally, STRAFE was found to outperform binary outcome algorithms in predicting fixed-time risk, possibly due to its ability to train on censored data. We show that STRAFE predictions can improve the positive predictive value of high-risk patients by 3-fold, demonstrating possible usage to improve targeting for intervention programs. Finally, we suggest a novel visualization approach to predictions on a per-patient basis. In conclusion, STRAFE is a cutting-edge time-to-event prediction algorithm that has the potential to enhance risk predictions in large claims datasets.

}




\maketitle

\newpage

\section{Introduction}\label{Introduction}

The use of big data in healthcare has been gaining increasing attention in recent years \cite{Vayena}. Longitudinal claims databases, which track an individual's health progress over time, are one type of big data that holds immense potential for improving patient care \cite{Raghupathi}. These databases contain a wealth of information, including diagnoses, procedures, medications, and lab tests, that can provide valuable insights into a patient's health journey. Claims databases are designed as sequences of visits, where each visit consists
of a collection of clinical information that was recorded at the same time. The number of visits and intervals between them can vary from patient to patient. Traditional machine learning models have used claims data for risk prediction, but they often treat each visit as an unordered collection of features and ignore the underlying graphical structure that reflects the physician's decision-making process.

In light of these challenges, the use of transformer-based models has shown promise in improving the ability to model time-series clinical data and predict clinical events. Transformers are based on the principle of self-attention for processing input data  \cite{Attention}. In contrast to traditional convolutional or recurrent layers, which are based on local interactions, self-attention allows the model to consider all input elements simultaneously when making predictions. Transformers are therefore ideally suited to tasks that require a global understanding of the input. The use of transformer-based architectures has been demonstrated to improve the ability to model the complexity of time-series clinical data by creating embedding representations for clinical notes \cite{Moerschbacher} and predict clinical events \cite{Zeng} \cite{Rao}. Kodialam et al. presented SARD, a transformer-based architecture that was used to construct a representation of patients based on claims information and to predict clinical outcomes \cite{SARD}. 

Chronic kidney disease (CKD) is one of the most widespread public health problems, affecting approximately 10\% of the world's population \cite{CHEN2016140} \cite{Subasi}. CKD is a progressive condition in which the kidneys gradually lose their ability to function properly. Stage 5 CKD, also known as end-stage renal disease (ESRD), is the final and most severe stage of the disease. At this stage, the kidneys have lost nearly all of their function and can no longer effectively remove waste and excess fluid from the body. Patients with stage 5 CKD require regular dialysis or a kidney transplant to sustain their lives. It is important for patients with CKD to receive early and consistent medical care to slow down the progression of the disease and prevent its progression to stage 5. Several studies have used large CKD datasets to predict deterioration, including using classic machine learning models \cite{Chen_2016} and neural networks \cite{Dutta}  \cite{Anupama} \cite{hamparia}, and even a transformer-based architecture \cite{TRACE}.

The implementations described above all focused on performing fixed-time risk prediction, which involves predicting the probability of an event occurring or not in a fixed-time frame. However, fixed-time risk prediction is often inappropriate for predicting clinical events due to the presence of censored data \cite{Luke}. Censored data refers to patients who have not yet experienced the event of interest and have limited observation time. In such cases, the observation time is not complete, and the event of interest has not yet occurred. This creates several issues when using fixed-time risk prediction for clinical event prediction. First, it can lead to a biased estimation of the event probability because only a portion of the patients has been observed for the entire observation time. Second, it can also result in a loss of information as the censored data provide limited information about the event. Third, it may also lead to over-optimistic predictions as the model is only trained on patients who have experienced the event and not on those who have not. To accurately predict clinical events, it is important to account for censored data and use appropriate methods such as survival analysis, which takes into consideration both the event and the censoring information.

Here, we present a novel architecture named STRAFE, a generalizable survival analysis transformer-based architecture for electronic health records, which utilizes the transformer computing power. STRAFE uses the Observational Medical Outcomes Partnership (OMOP) Common Data Model (CDM) standardized vocabularies, incorporates convolution of time-series visits and a self-attention mechanism to improve accuracy, and outputs a prediction for the time of the event. We demonstrate how STRAFE outperforms classical machine-learning models and other deep-learning architectures in predicting the exact time of deterioration to stage 5 CKD in a large retrospective claims dataset of over 130,000 individuals with stage 3 CKD. We also show how STRAFE can be used for fixed-time risk prediction and that training on censored data improves such predictions. Finally, we present a novel visualization to explain predictions on a per-patient basis and explore how STRAFE predictions can be used to improve care management interventions.

\section{Results}\label{Results}

\begin{figure*}
  \centering
  \includegraphics[width=1\textwidth]{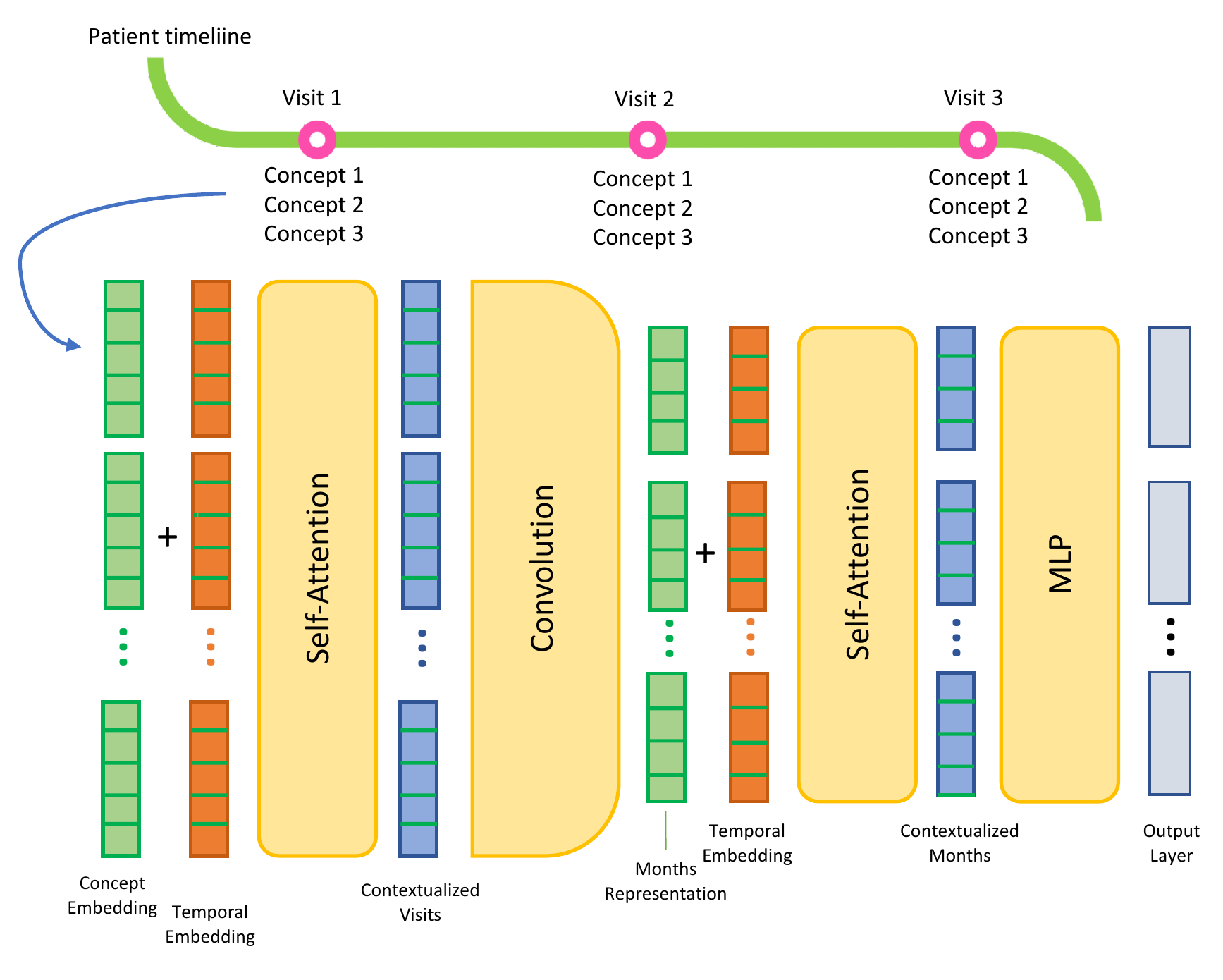}
  \caption{\textbf{STRAFE architecture.} Visits of patients consist of a set of concepts. The concepts of each visit are transformed into a concept embedding vector from a pre-trained embedding. These vectors, together with a temporal embedding vector, are fed into a self-attention layer which outputs contextualized visits. Using a convolution layer these are transformed into representations per month. A second self-attention mechanism is used to contextualize the months, and a MLP layer is used to extract survival probabilities per month. }
  \label{fig:architecture}
\end{figure*}

We developed STRAFE based on the architecture presented by Kodialam et al. \cite{SARD} (Figure \ref{fig:architecture}). The input data is a sequence of visits per individual in an OMOP CDM format. Each visit consists of a set of systematized nomenclature of medicine - clinical terms (SNOMED CT), which are transformed to embedding using pre-trained model trained on the full dataset. All embeddings of a visit are summed to create a vector for each visit, and an additional temporal embedding is provided as input. Context is provided to the visits by a self-attention layer. In STRAFE we then added a convolution layer that represents 48 months of risk prediction. A second self-attention layer and a multilayer perceptron (MLP) layer are utilized to calculate the probabilities of the occurrence of the event in each of the 48 months. For further details on the STRAFE architecture and implementation, refer to the \textbf{Methods} section.

\begin{figure*}
  \centering
  \includegraphics[width=1\textwidth]{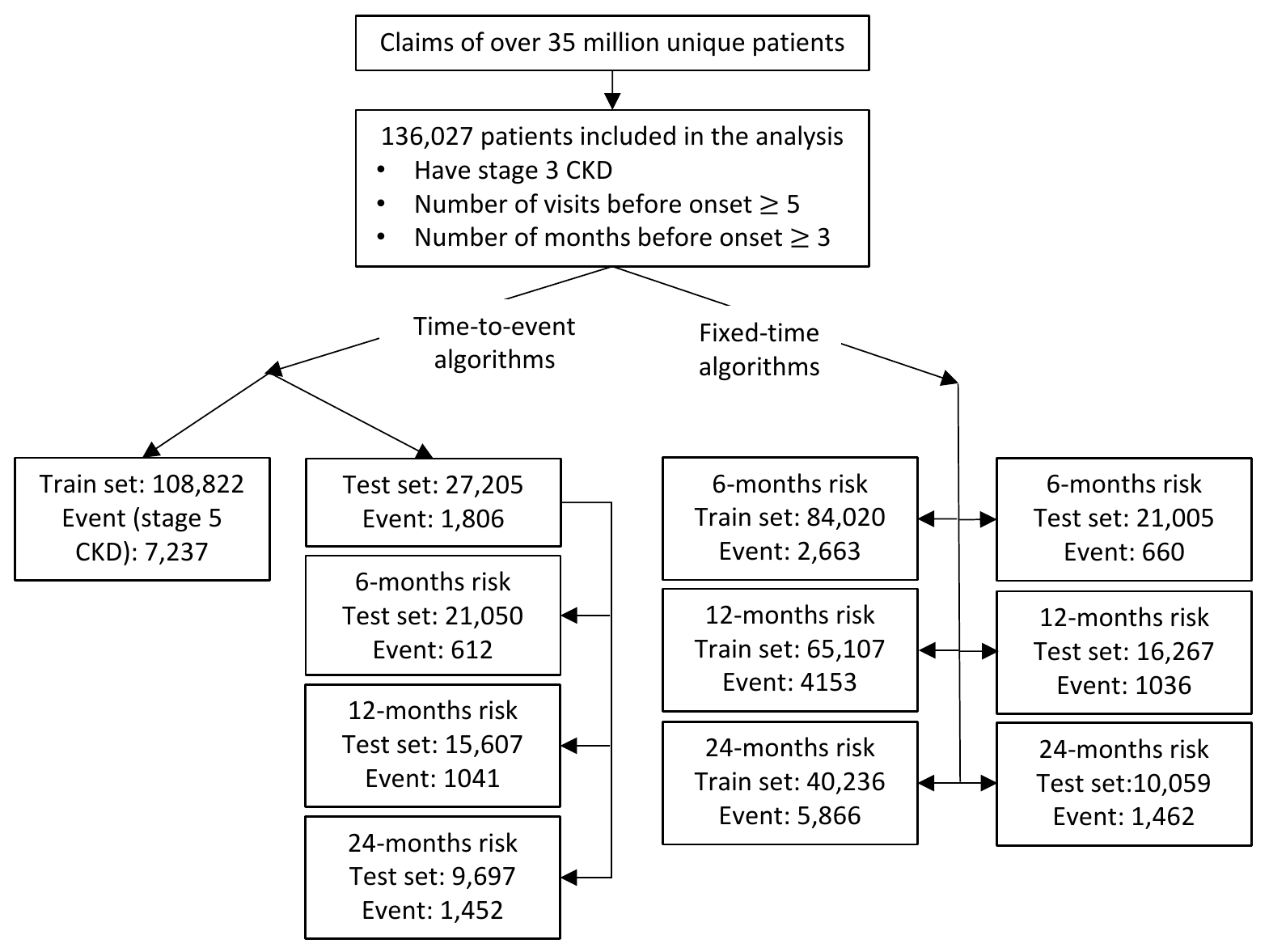}
  \caption{\textbf{Cohort selection procedure.} A de-identified claims database from a large US health insurance was used. A cohort of stage 3 CKD patients was identified and divided to train and test sets using 80/20 split. Time-to-event algorithms train on the full cohort. while fixed-time algorithms can only use non-censored patients.}
  \label{fig:cohort}
\end{figure*}

We devised a cohort of 136,027 patients with a code of stage 3 CKD, after the exclusion of patients without at least three months and five visits of data prior to the first indication (Figure \ref{fig:cohort}). This first indication was defined as the index date. The input data for each patient consists of basic demographic features (gender, age) and a sequence of visits with diagnoses, procedures, and pharmacy claims collected up to the index date (Figure \ref{fig:timetask}a). Following the index date, 8,896 patients ($6.64\%$) deteriorated to stage 5 CKD. The first indication of stage 5 code was defined as the outcome, and the time between the index date and the outcome date was defined as the time-to-event (or survival time). The dataset was divided into training and a held-out set at an 80/20 ratio (Table \ref{table:cohort}).

\begin{table*}[]
\small
\begin{tabular}{@{}ccc@{}}
\toprule
                                        & Training set (n=108,822) & Test set (n=27,205) \\ \midrule
Age, mean $\pm$ SD                        & 61 $\pm$ 13.37                & 61 $\pm$ 13.38           \\
Female                                  & 43.4\%                   & 43.5\%              \\
No. of visits, mean $\pm$ SD              & 52.33 $\pm$ 61.01               & 52.39 $\pm$ 61.56         \\
Stage 5 CKD                             & 6.66\%                   & 6.65\%              \\
Time-to-event, mean $\pm$ SD              & 558.12 $\pm$ 454.28               & 559.51 $\pm$ 452.37 \\
Charlson Comorbidity Index              & 4.79                     & 4.8                \\
Acute Myocardial   Infarction           & 6,293 (5.78\%)              & 1,602 (5.88\%)          \\
Congestive Heart   Failure              & 23,480 (21.57\%)          & 5,860 (21.54\%)      \\
Peripheral Vascular   Disease           & 24,154 (22.19\%)           & 6,152 (22.61\%)       \\
Cerebrovascular   Disease               & 19,183 (17.62\%)            & 4,719 (17.34\%)         \\
Dementia                                & 6,009 (5.52\%)            & 1,525 (5.61\%)         \\
Chronic Obstructive   Pulmonary Disease & 24,210 (22.24\%)            & 6,139 (22.56\%)         \\
Rheumatoid Disease                      & 6,530 (6.0\%)            & 1,622 (5.96\%)       \\
Peptic Ulcer Disease                    & 2,416 (2.22\%)              & 517 (1.9\%)          \\
Mild Liver Disease                      & 12,554 (11.53\%)            & 3,226 (11.85\%)         \\
Diabetes Without   Complications        & 3,426 (3.14\%)             & 865 (3.18\%)         \\
Diabetes With   Complications           & 39,408 (36.21\%)          & 9,945 (36.55\%)      \\
Hemiplegia or   Paraplegia              & 10,823 (9.94\%)            & 2,699 (9.92\%)     \\
Renal   Disease                         & 108,822 (100\%)          & 27,205 (100\%)     \\
Cancer (any   malignancy)               & 26,823 (24.64\%)              & 6,648 (24.43\%) \\
Moderate or Severe   Liver Disease      & 1,219 (1.1\%)            & 283 (1.0\%)         \\
Metastatic Solid Tumor                  & 3,175 (2.91\%)            & 857 (3.15\%)        \\
AIDS/HIV                                & 745 (0.68\%)             & 196 (0.7\%)         \\ \bottomrule
\end{tabular}
    \caption{Cohorts characteristics.}
    \label{table:cohort}
\end{table*}

\subsection{Time-to-event task} \label{sec:Time-to-event Task}

\begin{figure*}
  \centering
  \includegraphics[width=1\textwidth]{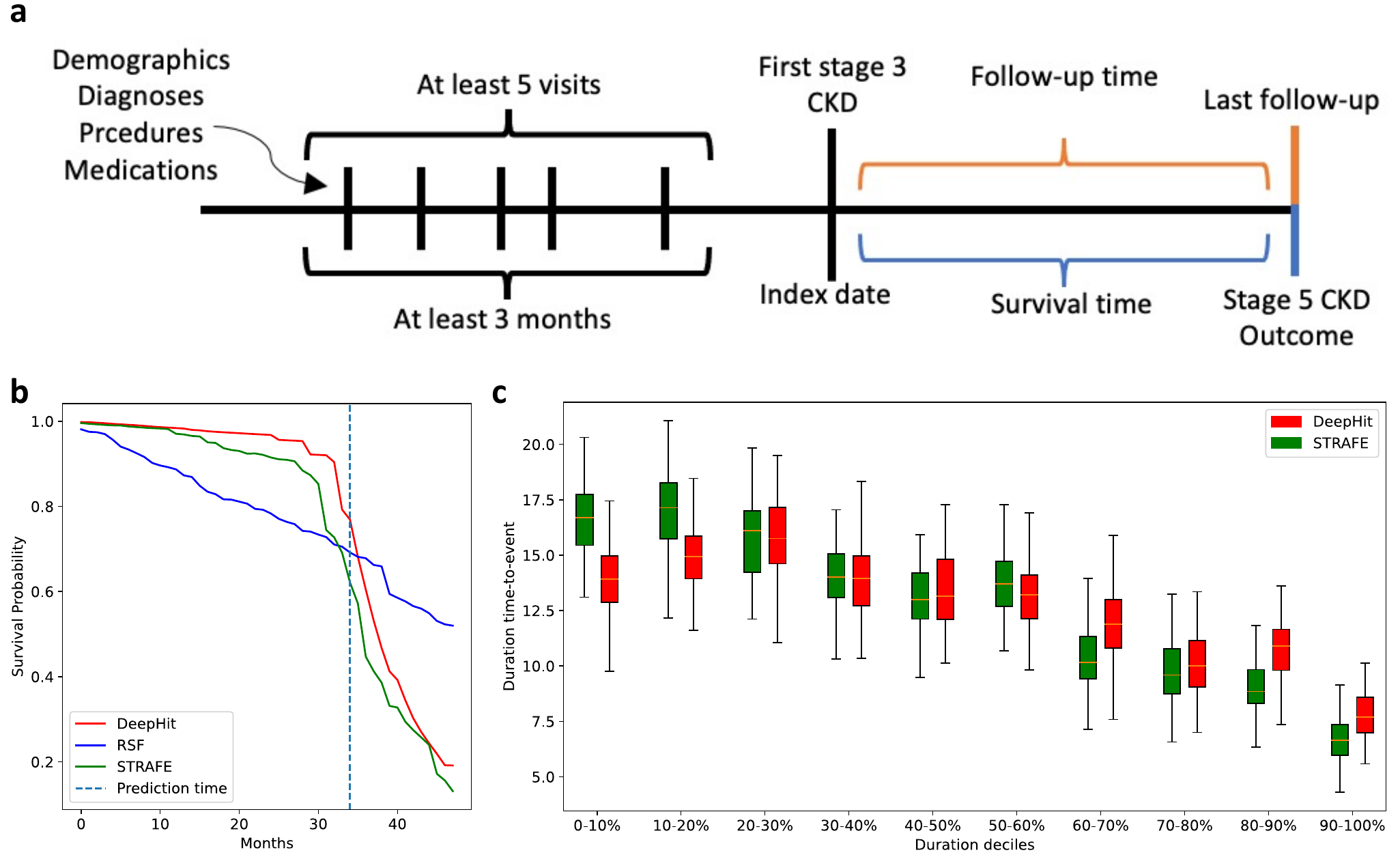}
  \caption{\textbf{Time-to-event prediction}. \textbf{a.} Study design: the observation period includes all visits prior to and including the first stage 3 CKD indication. Patients were then followed up until a stage 5 CKD indication or the last claim. \textbf{b.} Example of predicted survival curves for one patient who had an event after 34 months according to three algorithms. A survival curve that presents a sharp decline around the time of the actual event is considered an informative curve. \textbf{c.} Boxplots of duration time-to-event in each decile. Test set patients were grouped into deciles according to the predicted mean survival time.}
  \label{fig:timetask}
\end{figure*}

The STRAFE risk predictions are probabilities for the event in each month following the index date. We benchmarked STRAFE against several other survival analysis algorithms, including random survival forest (RSF) \cite{RSF} and DeepHit (a transformer-based survival analysis algorithm) \cite{Deephit}. To illustrate the challenges in survival analysis prediction we look at the survival curves obtained using STRAFE and the baseline algorithms for one patient, which deteriorated to stage 5 at 34 months after the index date (Figure \ref{fig:timetask}b). The survival curve is a monotonic decreasing function, which provides the probability of no event at each time point. In this example, STRAFE and STRAFE-LSTM predicted the highest risk at the time of the event (34 months). Importantly, in their prediction, and also of DeepHit, we observe that the sharp decline in the curve starts only several months before the time of the event, thus providing an accurate prediction of the time-to-event. The prediction of RSF (in this example) does not provide information on the time-to-event, as it declines monotonically along the time frame.

We trained each algorithm on the training set, and calculated both the C-index and mean average error (MAE) metrics on the held-out test set (Table \ref{table:cindex_and_me}). Since STRAFE uses embeddings, we used the embeddings as input for the other algorithms but also compared variations of bag-of-words (BOW) as input. Furthermore, to understand how the different layers in the architecture affect the prediction, we included three variations of STRAFE: replacing the second self-attention layer with a long short-term memory (LSTM) layer (STRAFE-LSTM) \cite{lstm}; removing the first self-attention layer, thus making the data uncontextualized (Uncont. STRAFE); and a combination those two (Uncont. LSTM). 

The C-index metric is a concordance metric of the predicted event time and the actual event time and is applied to all samples, including censored samples. The MAE metric on the other hand is applied only to samples that had an event during the follow-up period and is the average difference between the predicted event times and the actual event times. We first noticed that embedding had a major impact on the C-index predictions, however, had no effect on MAE. Next, all algorithms with embeddings had similar C-index values (except the uncontextualized STRAFE variant, which was a bit lower). However, STRAFE and its variation, all showed major improvement in the MAE values, suggesting, that while STRAFE does not improve rankings, it does provide significant value in predicting the actual time-to-event.

\begin{table*}[]
    \centering
    \begin{tabular}{ |p{4cm}|p{2cm}|p{2cm}|}
 \hline
 Model &  C-index & MAE  \\
 \hline  \hline
 RSF - BOW  &  0.6089  & 32.333   \\ \hline
 RSF - embeddings  &  \textbf{0.7187} & 31.853 \\ \hline \hline
 DeepHit - BOW   &  0.5799 & 28.391 \\ \hline
 DeepHit - embeddings   &  0.7144 & 28.59 \\ \hline \hline
 STRAFE   &   0.7101 & 22.164  \\ \hline
 STRAFE-LSTM   &  0.71 & \textbf{21.588} \\ \hline
 Uncont. STRAFE   &  0.6896 & 22.1427  \\ \hline
 Uncont. LSTM   &  0.7106 & 23.038 \\ \hline 
\end{tabular}
    \caption{\textbf{Time-to-event test evaluation results.} C-index and MAE values for the test set. BOW: bag-of-words.}
    \label{table:cindex_and_me}
\end{table*}

\par
 
To visualize the improved time-to-event prediction of STRAFE we performed a progression rate analysis. Samples with an event were ranked based on their predicted mean survival time (\textbf{Methods}) and were divided into deciles accordingly (Figure \ref{fig:timetask}c). Visualizing the actual event times in each decile we observed that the samples with the top deciles according to STRAFE indeed had shorter time-to-event, in accordance with MAE results. A similar analysis, where samples were ranked according to DeepHit, yielded lower performance, and less monotonic ordering of the deciles.

\par

\subsection{Fixed-time risk task}
\label{sec: risk task}

\begin{figure*}
  \centering
  \includegraphics[width=1\textwidth]{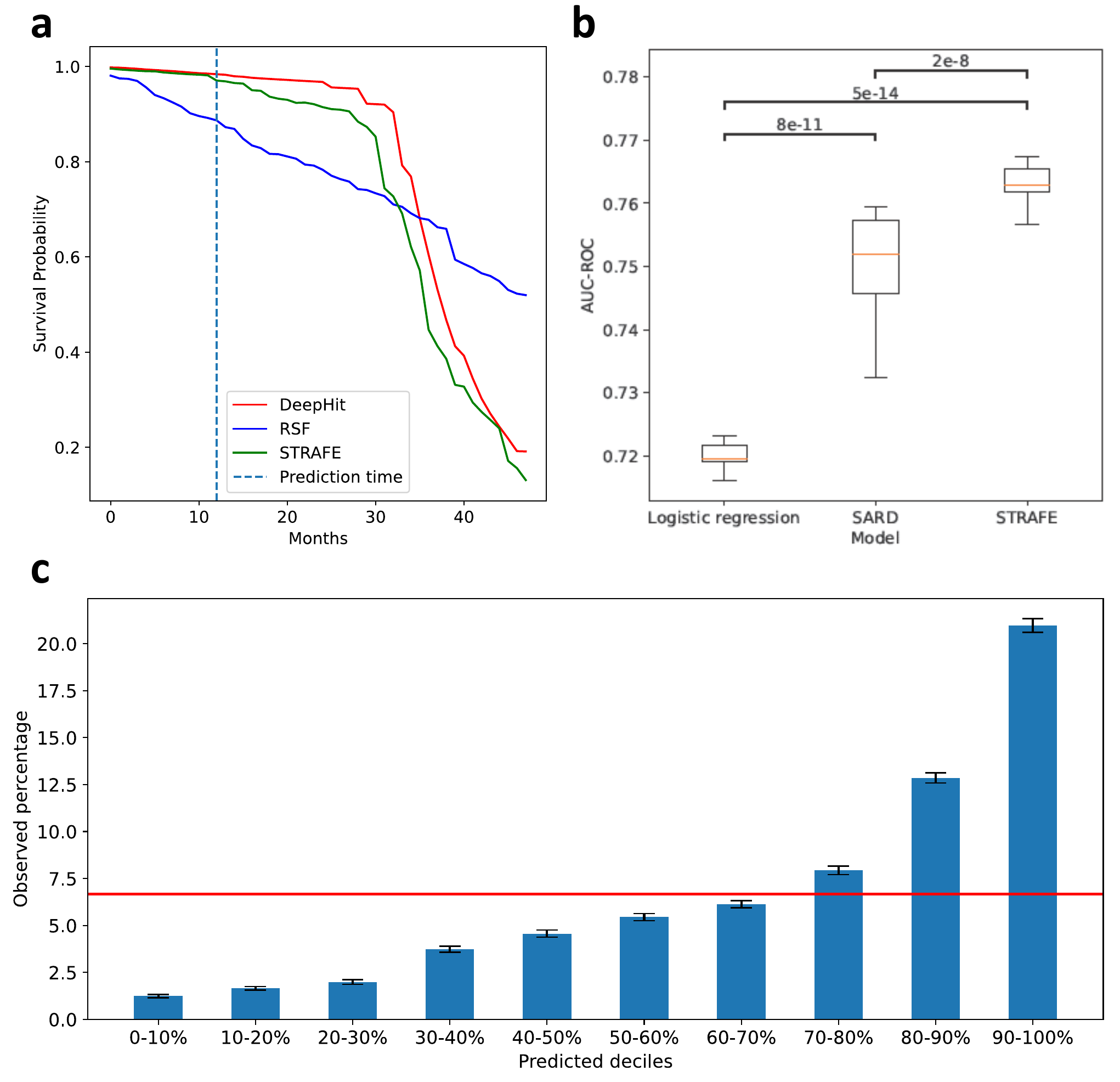}
  \caption{\textbf{Fixed-time prediction.} \textbf{a.} An illustration of transforming a time-to-event prediction to fixed time prediction - the probability at a fixed-time is used as the predicted probability. \textbf{b.} Boxplots of bootstrapped AUC-ROC values for three algorithms on the test set in the 24-months task. P-value(STRAFE vs. SARD) = 2e-8. \textbf{c.} Test set patients were divided into deciles according to the predicted probabilities of STRAFE at 12-months. Y\-axis shows the percentage of patients who deteriorated to stage 5 CKD. The red line is the percentage in the full test cohort.}
  \label{fig:figure4}
\end{figure*}

In our cohort, many patients had limited follow-up time, which is a common challenge in clinical datasets. Fixed-time risk algorithms require complete follow-up time, and as a result, many patients with short follow-up time must be omitted during training. In contrast, survival analysis algorithms can be trained on censored patients, including those with limited follow-up time. We investigated whether training with the full dataset and converting the predictions to fixed-time predictions at a specified time frame would result in improved accuracy compared to algorithms that can only be trained on non-censored samples (Figure \ref{fig:figure4}a). To this end, we compared the performance of STRAFE to logistic regression and the SARD algorithm, which STRAFE augments. We also compared the performance of other algorithms, converting their time-to-event predictions to fixed-time predictions for comparison purposes. We also added the algorithms used in the previous section and similarly converted their time-to-event prediction to fixed-time predictions.

Here, we used the area under the receiver operating curve (AUC-ROC) metric, and the analysis was performed for 6-, 12- and 24 months time frames (Table \ref{table:risk results}). First, similarly to the time-to-event task, we observed that embedding significantly improves performance. We also observed that the deep-learning algorithms performed slightly better than traditional machine-learning algorithms. Finally, STRAFE outperformed the other methods, including SARD, across all time frames (p-value[STRAFE vs. SARD] = 2e-8; Figure \ref{fig:figure4}b). We also noticed that the uncontextualized versions of STRAFE had similar performance to SARD, indicating the importance of attention in the representation phase.

The results of the comparison between STRAFE and other methods showed that STRAFE had superior prediction accuracy. However, both C-index and AUC-ROC metrics do not provide a clear path toward clinical usage. We aimed to demonstrate how STRAFE could be applied in a real-world clinical setting where early interventions are crucial for reducing the risk of disease progression. With this in mind, we simulated the usage of STRAFE by ordering patients based on their predicted risk of deteriorating to stage 5 within one year and dividing them into deciles. The fraction of patients who actually deteriorated was calculated for each decile, giving us a more concrete picture of STRAFE's potential impact in real-world applications (Figure \ref{fig:figure4}c). At the highest predicted risk group (top decile) the positive predictive value (PPV) was 20.9\%, which is more than three-fold higher than the full cohort (6.67 \%). Similarly, for 24-month risk, the PPV of the top decile was 28.42\% compared to 14.98\% in the full cohort. The analysis provides us with a possible clinical application that targets the highest-risk patients and provides them with a preventative treatment that is more likely to be successful than a random choice of patients.

Upon examining the predictive capacity of STRAFE, we discovered that its performance varied across different demographic groups. The AUC for the entire patient population was approximately 0.75; however, it rose to nearly 0.8 for individuals below the age of 60. Additionally, a marginally higher AUC was observed for males compared to females (0.761 vs. 0.748). Unfortunately, we lacked access to information on ethnicity, socioeconomic status, and geographical location. The discrepancies in predictive power may stem from variations in disease progression, biological traits, or other demographic attributes. In order to enhance the predictive accuracy of STRAFE for a broader range of patients, further research is required to pinpoint the specific factors driving these disparities and to fine-tune the model accordingly. By addressing these limitations, we can ultimately bolster patient outcomes and facilitate more informed healthcare decision-making for diverse populations.

\begin{table*}[]
    \centering
    \begin{tabular}{ |p{3cm}|p{1.5cm} p{1.5cm}  p{1.5cm}| }
 \hline
 Model & 6 months & 12 months & 24 months \\
 \hline  \hline
 LR - BOW  & 0.622 & 0.598 & 0.603 \\ \hline
 LR - embeddings  & 0.711 & 0.710 & 0.720 \\ \hline \hline
 SARD   & 0.725 & 0.731 & 0.748\\ \hline \hline
 RSF - BOW  & 0.628 & 0.609 & 0.578 \\ \hline
 RSF - embeddings  & 0.719 & 0.723 & 0.683\\ \hline \hline
 DeepHit - BOW   &  0.587 & 0.576 & 0.586\\ \hline
 DeepHit - embeddings   & 0.729 & 0.728 & 0.725\\ \hline \hline
 STRAFE   &  0.751 & 0.754 & $\mathbf{0.764}$ \\ \hline
 STRAFE-LSTM   & $\mathbf{0.754}$ & $\mathbf{0.756}$ & 0.763\\ \hline
 Uncont. STRAFE   & 0.719 & 0.719 & 0.738 \\ \hline
 Uncont. LSTM   & 0.742 & 0.746 & 0.756\\ \hline 

\end{tabular}
    \caption{Fixed-time task evaluation metrics. AUC-ROC values for test sets of fully observed patients at 6-, 12- and 24-months.}
    \label{table:risk results}
\end{table*}
\par

\subsection{Per-patient explainability} \label{sec:Attention Analysis results}

The self-attention mechanism provides temporal and structural context to visits, and  deeper and more complex relationships can be revealed between visits in a patient's medical history. However, it is difficult to develop a comprehensive explainability method for all patients that refers to specific features in relation to the entire dataset, as is done in classical machine learning. Nevertheless, it is possible to examine individual patients and determine which visits are most critical, which may be helpful in determining the primary reason for their elevated risk. The input to the self-attention mechanism is a sequence of visits, not a diagnosis, so we can only analyze attention weights based on visits.

\begin{figure*}
  \centering
  \includegraphics[width=1\textwidth]{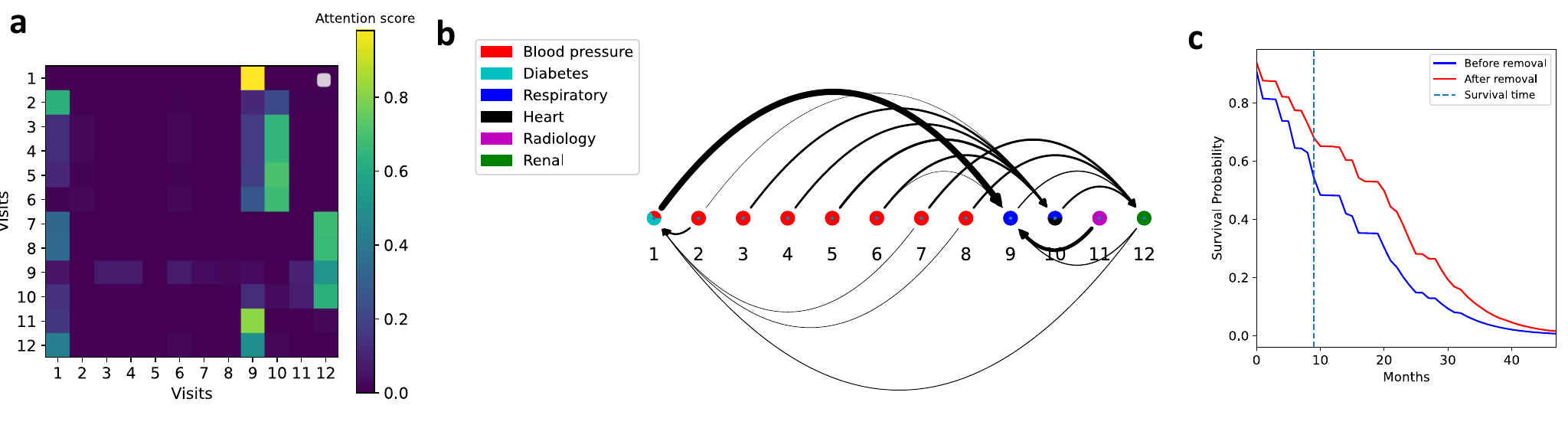}
 
  \caption{\textbf{Per-patient explainability.} \textbf{a.} An example heatmap of the interaction values of the attention matrix for a patient that had 12 visits. Visit 1 and 9 have the highest value. \textbf{b.} A graph-based approach for visualizing the connectivity between the visits. Dots were colored according to ICD-10 code chapter of the primary diagnosis of the visit. \textbf{c.} The original predicted survival curve for the patient compared to the predicted survival curve following the removal of visits 1 and 9.}
   \label{fig:figure5}
\end{figure*}

By analyzing the attention weights in the representation phase, we can identify the most closely related visits in the patient's history. Those visits that are most connected are those that have the highest attention scores between them. It should be noted, however, that the self-attention mechanism captures non-directional relationships between visits (i.e., they do not cause one another). Domain experts can gain insight into how each visit is contextualized by examining attention weights in STRAFE's representation phase. The physician may obtain the predicted time to deterioration to stage 5, and then examine the reasons behind the model's decision for the prediction by examining the most important visits. In order to demonstrate the clinical utility of this analysis, we selected a patient diagnosed with stage 5 nine months after the onset of stage 3 (Figure \ref{fig:figure5}a-b).

Using this representation it is possible to detect the inner interactions in the medical history of the patient in relation to the predicted event. Here, we observed strong connections between the former blood pressure visits and the respiratory visits. Additionally, the 9th visit was found to be a key factor in the patient's rapid progression, with connections to several other visits. Diabetes and respiratory events may have played a role in the rapid progression of this patient's illness. Although some connections can be inferred from this graph by domain experts, there are still some unknown characteristics, such as how to interpret the edge direction. To demonstrate the impact of these strong connections, we conducted an experiment where the visits with the highest connections, 1 and 9, were omitted  (Figure \ref{fig:figure5}c). The result was a noticeable shift in the predicted time of deterioration, showing the importance of these connections in STRAFE's predictions. This approach offers a unique way to visualize and analyze temporal clinical data.

\section{Discussion}\label{Discussion}

We presented STRAFE, a novel architecture we developed for modeling time-series clinical data and predicting time-to-event. In order to predict patient deterioration in CKD, the STRAFE algorithm was applied to real-world claims data in the OMOP CDM format. We evaluated STRAFE's performance by comparing it to other survival analysis algorithms and found that it outperforms other methods in the accuracy of the time of the event. We further showed that STRAFE can perform improved predictions also for fixed time-frames.

While STRAFE outperformed other algorithms in the accuracy of the exact event time, it did not provide improvement according to the C-index metric. It should be noted that the C-index metric has some disadvantages. First, it is highly sensitive to class imbalance. This can lead to a biased evaluation of the prediction performance, particularly for rare events \cite{Concordance_Probability}. Second, the C-index only measures the ranking performance of the predictions, but it does not consider whether the predicted probabilities are well-calibrated, which is essential for decision-making and clinical use. Therefore, the combination of C-index and MAE is a better approach for evaluating machine-learning methods in survival analysis.

While the most appropriate way to analyze clinical events is with survival analysis, however, fixed-time risk prediction is yet the most common approach in the literature. Fixed-time risk prediction requires excluding all censored data, which may cause multiple types of biases, but also just causes a significant loss of training data. Since our time-to-event method does not lose this information, we hypothesized that the inclusion of the censored data will enhance performance. Interestingly, this proved correct: when applied to a fixed-time task, STRAFE outperformed the baseline methods, including SARD which STRAFE was built upon its architecture. Similarly, RSF outperformed LR suggesting that censored data improves performance in a more general sense and not a specific advantage of STRAFE.

We provided here a new form of explainability of the attention mechanism appropriate for clinical time-series data. Inspection of the STRAFE model's self-attention mechanism enabled a nuanced understanding of which medical timeline elements are most relevant for prediction or are related to one another based on contextual information from the patient's medical history. Thus, if it is necessary to interpret an entire clinical narrative, a deep model will be more accurate than simpler baselines. Additionally, it may illustrate another benefit of using a self-attention-based model, even if the results were not satisfactory. In clinical research, good results are of limited value without a comprehensive explanation. We proposed a novel approach where the physician will be able to identify which visits are more closely related to each other and to the anticipated event. It should be noted that this method is not without limitations. Our first concern is that the scale of attention scores is not yet clear. Based on the scores we have, we cannot draw any conclusions regarding how closely connected the two visits are. Additionally, it is possible that the attention weights do not provide a complete picture, since other components of the architecture, such as convolutional weights and fully connected layers, also influence the final result. Moreover, the connections may reflect a correlation rather than a causal relationship, and therefore, the results may differ between different types of patients. Because the visit representation consists of all concepts within a visit, we cannot conclude with high confidence that the concepts within specific visits are related or cause each other. However, domain experts can be used to identify the dominant concept in each visit.

Our implementation of STRAFE can be further improved to enhance performance. STRAFE has the disadvantage that it aims to minimize loss across all time phases, thereby complicating the training process. Consequently, further investigation of the architecture may result in improved results. For example, we did not observe a difference in the performance between using self-attention or LSTM at the prediction phase; however, we did observe an improvement compared to the uncontextualized variations. Thus, it appears that attention plays a greater role during the representation phase than during the prediction phase. As a result of the use of attention in the representation of the input sequence, the model can obtain contextual information and recognize the relationships between different visits.

In addition to the architecture, it is possible that a poor choice of hyperparameters may have been made during the optimization process, such as the optimizer and learning rate, or that some useful techniques for optimizing neural networks, such as weight initialization and momentum, would have improved performance.

Finally, we demonstrated possible clinical usage for STRAFE by identifying the highest-risk patients. The ability to accurately identify patients at high risk and prioritize their needs can result in improved health outcomes, reduced costs, and more efficient use of resources. Furthermore, this approach can be easily scaled and adapted to other diseases and healthcare settings, making it a valuable tool for healthcare providers and payers. In conclusion, we believe that STRAFE has the potential to significantly impact patient care by enabling more targeted and effective interventions.


\section{Methods}\label{Methods}

\subsection{STRAFE architecture}

Claims data provide a wealth of clinical information regarding diagnoses, procedures, prescriptions, and so on. As discussed earlier, one of the main challenges when using claims data is how to represent the information it contains. The irregularity of time and the differences in type and amount of information between events require a gentle and sophisticated way to extract the information from the raw data. 
Kodiolam et al. tried to face this challenge by introducing a Transformer-based architecture called SARD \cite{SARD}. In order to obtain a contextual representation, we used the same embedding creation process which is based on the word2vec method \cite{Mikolov}, and then fed it into the self-attention mechanism. SARD uses this representation to estimate the probability of the target event. Our work aims to predict time-to-event, so we fed the contextualized representation into time-to-event head (Figure \ref{fig:architecture}).

In order to describe the architecture we will start with some notation.
We denote the set of visits made by a patient $i$ by $V_i$ and represent this patient’s $j$-th visit by $V_j^i$. 
We further denote the time of visit $V_j^i$ by $t_j^i$ and the set of codes assigned during visit $V_j^i$ with $C_j^i \subseteq C$.
The vector representation $\psi(V_j^i) \in \mathbb{R^{d_e}}$ of each visit is calculated as:
\begin{center}
\begin{equation}
\abovedisplayskip=0pt
\psi(V_j^i) = \sum_{c \in C_j^i} \phi(c)
\end{equation}
\end{center}
providing invariance to permutations of the codes.
\par The sequence of events is not explicitly encoded, and visits do not occur at fixed intervals. Therefore, the time of each visit is embedded using sinusoidal embeddings \cite{Attention}:
\begin{center}
\begin{equation}
\abovedisplayskip=0pt
\tau(V_j^i) = sin(\tilde{t_j^i}\omega) \mathbin\Vert cos(\tilde{t_j^i} \omega)
\end{equation}
\end{center}

We denote concatenation with $\mathbin\Vert$, $\omega$
is a length $d_e / 2$ vector of frequencies in geometric progression from
$10^{-5}$ to 1, and sin and cos are applied element-wise.
\par After the embedding phase, the claims information of a single patient is represented with a number of vectors which is a hyperparameter which is written as $n_v$ which limits the number of visits we use to represent the patient. Now the embeddings can be fed into the Transformer-based architecture. First, $\psi(V_j^i)$ and $\tau(V_j^i)$ are summed to create final encodings that represent the content and timing of visits. To contextualize visits in a patient’s overall history, multi-headed self-attention with $L$ self-attention blocks and $H$ heads are used.
For efficiency, only the $n_v$  most recent visits are used and padding for patients with less than $n_v$ visits is applied. 
\par In addition, dropout with probability $\rho_t^d$ is applied after each self-attention block to prevent overfitting. This
approach allows any visit to attend to any other, so longer-range dependencies of clinical interest can be learned. 
The contextualized embedding of visit $V_j^i$ is found by utilizing the self-attention mechanism.
This process is then repeated at each attention layer using the contextualized embeddings as inputs, and residual connections are used between layers. The outputs of each head are concatenated to create final, contextualized visit representations $\psi(V_j^i)$.

We can estimate the survival function using the Kaplan-Meier estimator \cite{kaplan_meier}. Thus, our model needs to predict the complement event of the hazard function: $q(t\mid X) = 1-\lambda(t\mid X)$. This should be done for all $t= 0,1,...,T_{max}$.
\par In this phase, we use the SARD representation (as derived from taking the self-attention outputs) of a patient, which consists of visits. In order to feed this patient representation into the survival analysis head  we use a convolutional layer that maps the current representation (matrix of shape $n_v \times d_e)$ to a matrix of shape  $T_{max} \times d_e$.
This new representation is directly fed into the survival head which contains another temporal embedding layer that aims to provide time awareness to each timestep.
The model outputs $T_{max}$ probabilities which are compatible with $q(t\mid X)$ for $t=0,1,..T_{max}$.

Our survival function estimation is based on the same principles as \cite{transformer_based_survival_analysis} which takes these outputs and then multiplies them in a recursive manner:
\begin{center}
\begin{equation}
\abovedisplayskip=0pt
S(t\mid X) = \prod_{\rho=0}^t q(\tau\mid X)
\end{equation}
\end{center}

The Mean survival time can be estimated by:  
\begin{center}
 \begin{equation}
 \mu = \sum_{t=0}^{T_{max}} S(t\mid X)
 \end{equation}
\end{center}

The estimation of the survival function is then fed into the loss function defined as:
\begin{center}
\begin{equation}
\abovedisplayskip=0pt
\mathcal{L} = \sum_{i\in observed} \mathcal{L}_{X_i}^{obs}  +  \sum_{i\in censored} \mathcal{L}_{X_i}^{cens}
\end{equation}
\end{center}

where the observed part is:

\begin{center}
\begin{equation}
\begin{split}
\abovedisplayskip=0pt
\mathcal{L}_X^{obs} = - \sum_{t=0}^{T-1} log \hat{S}(t\mid X)  - \sum_{t=T}^{T_{max}} log (1-\hat{S}(t\mid X))
\end{split}
\end{equation}
\end{center}

and the censored part is:
\begin{center}
\begin{equation}
\abovedisplayskip=0pt
\mathcal{L}_X^{cen} = - \sum_{t=0}^{T-1} log \hat{S}(t\mid X)
\end{equation}
\end{center}

The STRAFE architecture is a combination of all the blocks discussed above STRAFE (Figure \ref{fig:architecture}).

\subsection{Survival analysis}\label{Survival analysis}

Survival analysis, whether classic or deep learning-based, often assumes proportional hazards, a belief that is rarely evaluated and often violated. In order to overcome this assumption, we used a non-parametric discrete-time model in which the follow-up time is divided into time windows, each with its own hazard, and the model learns survival in all time windows simultaneously. Right-censored patients were handled by applying a tailored loss function, i.e., patients who did not experience the event within the observation period, either because they left the insurance company or the follow-up period expired in 2020. When the risk period extends beyond observation, time-to-event analyses are appropriate. In survival analysis, the training data consists of the features and time pairs $(X_i, T_i)$, where $T_i$ can be observed or censored (we consider only right censoring). Patients who have been diagnosed with CKD stage 5 are considered observed, and survival time is calculated as the time between diagnosis of stage 3 and diagnosis of stage 5. The survival time in all other cases is determined by the time between the onset of stage 3 and the last observation recorded in their records.

\subsection{Estimating risk using survival function}

We can transform the survival function estimator to predict the risk of an event occurring in a specific period of time. We define the fixed-time risk task to predict if an event will happen until time $T_R$. The meaning of the term $S(t)$ is the probability of a patient that the event will happen after the time $t$. Thus, the meaning of $1-S(t)$ is the probability that the event will occur until the time t, which is exactly what we are looking for in the risk task. We use the value of $1-S(T_R)$ as the probability given to a patient to have an event until the time $T_R$. We can use this probability to evaluate the ability of the model (which is trying to minimize survival loss) to succeed in the fixed-time prediction of the occurrence of an event.

\subsection{Data and variables}\label{Data and variables}

The Elevance Health Digital Data Sandbox powered by Carelon Digital Platforms was used for the experiments. This certified de-identified database contains information on over 35 million members between 2015 and 2020, and includes billing practices, demographics, diagnosis codes, procedures codes, and pharmacy claims. We used the convenient structure of the OMOP CDM \cite{ohdsi2019book} to perform efficient pre-processing and extracting applicable medical information for feeding the machine learning models.
In this database, diagnoses are represented by SNOMED CT codes, procedures by CPT codes, and prescriptions by RX norm. 

On the basis of the entire database of patients, we devised a cohort of patients who met the following inclusion criteria: they had been diagnosed with stage 3 CKD at least three months following their first visit, and they had had five visits prior to the diagnosis (Table 1; Figure \ref{fig:cohort}). To predict risk in a fixed time window (6-, 12- or 24 months), all patients censored before that time period were excluded. We divided our dataset into training and test sets at an 80/20 ratio, using stratification to maintain the class ratio between the training and test sets.

\subsection{Model development}

This section provides details regarding the implementation of the three central components in STRAFE. Concept embedding is the first step in our training process. Our work included feeding our models with pre-trained embeddings, which were trained on all the claims database (not only on the cohort) so that a greater variety of concepts could be learned. These embeddings can also be used to predict other clinical events in the future without the need for task-specific training. For the training of the concept embeddings, we utilized the Gensim package \cite{gensim}. There were 36,480 clinical concepts in our data, including 16,676 condition concepts, 10,378 procedure concepts, and 9,426 drug concepts. In order to apply the skip-gram method to the model, we created a "sentence" structure that is similar to that used in natural language processing (NLP). In this context, a "sentence" is a collection of all the codes that occur during a specific period of time. We selected a 90-day interval. Therefore, each "sentence" consists of all the codes present during a patient's 90-day period. The embedding size for the data was set to 128 as this is a common heuristic. The number of visits to a patient was limited to 100.

The second step is the self-attention mechanism. We performed a grid search in order to determine the optimal hyperparameters for the self-attention mechanism. Tests were conducted on the following combinations of hyperparameters:
\begin{itemize}
    \item Attention heads : \{1,2,4,8\}
    \item Number of stacked components: \{1,2,4,8\}
    \item Dropout : \{0.1,0.2,0.3, 0.5\}
\end{itemize}
The hyperparameters selected are four attention heads with only one component and dropout of 0.3. After obtaining the contextualized representation of visits from the self-attention mechanism, we passed the sequence into a convolution layer that sets the sequence length to the maximum time (in months) for which survival analysis should be conducted. Therefore, we performed a time-to-event prediction for 48 months following the onset of stage 3 CKD. In order to change the sequence length without making too complex transformations we used a convolution layer. This method proved to be the most effective after implementing other dimensional reduction methods, such as fully connected, which led to overfitting during training.
We then use another self-attention layer to contextualize the new sequence (with the addition of another temporal embedding).
In this component, we examined the same set of hyperparameters as well as whether increasing the element dimension (which was 128 in the first component) would improve the results. According to the results, there has been no significant improvement.

The final component is the creation of an output that represents the survival function. We used a batch size of 256 and the ADAM optimizer \cite{Adam} with a learning rate of 2e-3. In addition, weight decay and momentum were also tested, but neither of these methods improved the training process.

\subsection{Experiments}

We used ablation studies to empirically validate our design decisions. Among the components of STRAFE, the first is considered to be the representation phase, while the second is considered to be the prediction phase. Therefore, we propose three variants of the STRAFE architecture. We changed the prediction phase in the first variation. The second self-attention mechanism was replaced with an LSTM, which means the output of the convolution layer was fed to the LSTM head. As part of this variation, we tested the added value of using attention during the prediction phase. In the second variation of STRAFE, attention is eliminated from the representation phase. In this manner, the summed representation of the visits is fed directly into the convolution layer. Through this variation, we were able to determine whether using attention in representation adds any significant value.
Due to the fact that the self-attention mechanism is the first interaction between visits in the model architecture, we hypothesized that it plays a greater role in the representation phase. Thus, we anticipated improved results when the self-attention mechanism was applied. In spite of this, we did not anticipate significant improvements when using self-attention in the prediction phase, since we already have a contextualized representation from the previous phase, and the only structural change is the convolution layer. STRAFE is the main architecture, while STRAFE-LSTM is the first variation. The other variation will be referred to as uncontextualized-STRAFE and uncontextualized-LSTM.

We performed the experiments on a workspace with Tesla K80 and a
Ubuntu 18.04.3 LTS operating system. The neural networks were implemented using PyTorch 3.6 and the survival baselines were implemented using Scikit-survival and Pycox.

\subsection{Baselines}\label{Baselines}
The baselines used to compare with STRAFE are divided into two categories: baselines for time-to-event prediction and baselines for fixed-time risk prediction. Random survival forest (RSF) \cite{RSF} and DeepHit \cite{Deephit} were used as baselines for time-to-event prediction. We run RSF with 1000 estimators by using scikit-surv package  \cite{sksurv}. We run the DeepHit model which uses a softmax classifier to predict the survival distribution. Accordingly, we searched for the optimal hyperparameters in the following spaces:
\begin{itemize}
    \item num layers: \{1, 2, 4\}
    \item node size: \{32, 128, 512\}
    \item dropout: \{0.1, 0.3, 0.5\}
    \item alpha: \{0.1, 0.3, 0.5, 0.7, 0.9\}
    \item sigma: \{0.1, 1, 10\}
\end{itemize}

Optimal hyper-parameters are as follows:
num layers=3, node size=[512,128,32], dropout=0.5, alpha=0.5, sigma=0.1 with batch size of 256.

\begin{itemize}
    \item num layers: $\{1, 2, 4\}$
    \item node size: $\{32, 128, 512\}$
    \item dropout: $\{0.1, 0.3, 0.5\}$
    \item alpha: $\{0.1, 0.3, 0.5, 0.7, 0.9\}$
    \item sigma: $\{0.1, 1, 10\}$
\end{itemize}
Optimal hyper-parameters are as follows:
num layers=3, node size=[512,128,32], dropout=0.5, alpha=0.5, sigma=0.1 with batch size of 256.

We used two types of baselines for the risk task, an indirect baseline and a direct baseline. As indirect baselines, we used all models that have been trained for survival analysis, including RSF, DeepHit, and STRAFE variations. As direct baselines, we used SARD and logistic regression (LR).

We applied RSF, DeepHit, and LR using two different types of input: embedding-based approach, and a bag-of-words (BOW) approach. The embedded representation approach consists of summing up all embedded representations of all visits in order to obtain a vector representing the entire history of patient i: $\sum_{j} \psi (V_j^i)$. The summed vector will have a dimension of $d_e$. In the second approach, occurrences of each concept were counted within a fixed-length window of the patient's history. We chose to use the entire patient history as a time interval. This approach produces a large number of features (number of concepts), but also has a sparse representation.
In order to better understand the strengths of the concept embedding approach over the naive approach of a BOW, both input representations were used.

\subsection{Evaluation}\label{Evaluation}
In survival analysis, the C-index is a widely used metric \cite{cindex}. Rank is defined as the ratio of concordant pairs divided by comparable pairs. 
\begin{center}
\begin{equation}
     C-index = \frac{\sum_{i,j} \mathbf{1}[T_i<T_j]\cdot \mathbf{1}[\hat{T_i}<\hat{T_j}]\cdot \delta_i }{\sum_{i,j} \mathbf{1}[T_i<T_j]\cdot\delta_i }
\end{equation}
\end{center}
where $\delta_i = 1$ if $T_i$ is observed and 0 otherwise.
\par

As a result, if we order the patients by their predicted survival time and this is the same order as the real survival time, we obtain a score of 1. However, we can achieve that score even if we mispredict the exact survival time for each patient.
In spite of its simplicity, Harrell's concordance index has two main drawbacks:
\begin{enumerate}
    \item  When censoring increases, it becomes overly optimistic \cite{cindex}.
    \item The C-index does not evaluate the exact survival durations, so an inaccurate model can still score well.
\end{enumerate}

\par
In addition, we propose to evaluate precise duration predictions on observed subjects using mean absolute error (MAE):
\begin{center}
\begin{equation}
    MAE = \frac{\sum_{i} \mid T_i-\hat{T_i}\mid \cdot \delta_i}{\sum_{i} \delta_i}
\end{equation}
\end{center}


To estimate the model's performance on the fixed-time risk task, we used the area under the receiver-operator curve (AUC-ROC), which is the area under the plot of true positives versus false positives. The equivalent interpretation of a model is the probability that a random patient with a positive outcome would receive a higher score than a random patient with a negative outcome. While metrics such as accuracy can be useful in cases of class-balanced problems, and metrics such as AUC-PRC can also be useful in cases of extreme class imbalances, our metric is applicable to a wide range of class imbalances in clinical settings.

\subsection{Data availability}\label{dataavailability}

Due to confidentiality and contractual requirements, supporting data cannot be made openly available. Additional information about the data used in this research, and requests to access it, can be made by application to the Elevance Health Digital Data Sandbox powered by Carelon Digital Platforms Manager (datasandbox@carelon.com) at Carelon Digital Platforms (\url{https://www.carelondigitalplatforms.com/digital-data-sandbox}).

\subsection{Code availability}\label{codeavailability}

The code developed in this study is available at \url{https://github.com/dviraran/STRAFE}. STRAFE takes a large portion of its architecture from SARD, which is available at \url{https://github.com/clinicalml/omop-learn}.

\subsection{Ethics}\label{ethics}

This study was designed as an analysis based on medical claims data, and there was no active enrollment or active follow-up of study subjects, and no data were collected directly from individuals. The study was not required to obtain additional IRB approval, as the HIPAA Privacy Rule permits protected health information (PHI) in a limited data set to be used or disclosed for research, without individual authorization, if certain criteria are met.

\section{Authors Contribution}

The authors confirm contribution to the paper as follows: study conception and design: DA; algorithm development: MZ; analysis and interpretation of results: MZ \& DA; draft manuscript preparation: MZ \& DA. Both authors reviewed the results and approved the final version of the manuscript.

\section{Acknowledgments}

DA is supported by the Azrieli Faculty Fellowship. 

\section{Declarations}

DA reports consulting fees from Carelon Digital Platforms. This project was not performed as part of the consulting.

\bibliography{general.bib}
\end{document}